# Arabic Sign Language Recognition using Multimodal Approach

Ghadeer Alanazi

King Saud University, Riyadh, KSA, 445920679@student.ksu.edu.sa

Abir Benabid

King Saud University, Riyadh, KSA, abbenabid@ksu.edu.sa

Arabic Sign Language (ArSL) is an essential communication method for individuals in the Deaf and Hard-of-Hearing community. However, existing recognition systems face significant challenges due to their reliance on single sensor approaches like Leap Motion or RGB cameras. These systems struggle with limitations such as inadequate tracking of complex hand orientations and imprecise recognition of 3D hand movements. This research paper aims to investigate the potential of a multimodal approach that combines Leap Motion and RGB camera data to explore the feasibility of recognition of ArSL. The system architecture includes two parallel subnetworks: a custom dense neural network for Leap Motion data, incorporating dropout and L2 regularization, and an image subnetwork based on a fine-tuned VGG16 model enhanced with data augmentation techniques. Feature representations from both modalities are concatenated in a fusion model and passed through fully connected layers, with final classification performed via SoftMax activation to analyze spatial and temporal features of hand gestures. The system was evaluated on a custom dataset comprising 18 ArSL words, of which 13 were correctly recognized, yielding an overall accuracy of 78%. These results offer preliminary insights into the viability of multimodal fusion for sign language recognition and highlight areas for further optimization and dataset expansion.

## 1 INTRODUCTION

Human communication predominantly relies on natural language, which serves as a vital tool for interaction and social engagement. For individuals with hearing impairments (HI), including those who are deaf or hard of hearing (D/HoH), sign language serves as a critical mode of communication. Sign language constitutes a unique form of communication that utilizes gestures and body movements to convey meaning. Sign language recognition plays a crucial role in bridging the communication gap between the deaf community and society at large.

Sign language is a visual form of communication used by deaf and hard-of-hearing communities worldwide. It is a rich and complex language with its own grammar, vocabulary, and cultural nuances, distinct from spoken languages. While there is no single universal sign language, many different sign languages exist globally, each with its own unique characteristics, such as American Sign Language (ASL), British Sign Language (BSL), and Arabic Sign Language (ArSL). Developing automated sign language recognition (SLR) systems is a significant challenge due to the complexity of sign languages, including variations in signing styles, hand shapes, movements, facial expressions, and body posture. However, successful SLR systems have the potential to bridge communication gaps and significantly improve the lives of deaf and hard-of-hearing individuals. The development of effective SLR systems has been a focus of research for several decades. Early approaches relied on traditional computer vision techniques, but with the advent of deep learning, significant progress has been made. These advancements have not only improved recognition accuracy but have also opened up new possibilities for applications across various domains.

**Arabic Sign Language (ArSL)** is the primary form of communication used by deaf communities across Arab countries. Developing systems capable of recognizing ArSL gestures is essential to facilitating communication and supporting the integration of the Deaf into society. According to the World Health Organization (WHO) [**34**], approximately 5% of the global population experiences hearing impairment, highlighting the importance of accessible communication solutions for this significant demographic. The ability to automatically recognize ArSL can have profound implications for various aspects of society:

- **Education:** ArSL recognition can facilitate inclusive education by providing real-time translation of classroom lectures and educational materials, making them accessible to deaf and hard-of-hearing students. This can significantly improve their educational outcomes and opportunities.
- **Healthcare:** In healthcare settings, ArSL recognition can enable effective communication between deaf patients and healthcare professionals, ensuring accurate diagnoses, treatment plans, and overall better healthcare experiences.
- **Workplace Accessibility:** By providing automated translation in workplace environments, ArSL recognition can break down communication barriers and create more inclusive employment opportunities for deaf individuals.
- **Social Inclusion:** Beyond these specific domains, ArSL recognition can promote broader social inclusion by facilitating communication in everyday interactions, such as public services, social events, and online platforms.

While there have been many advancements in natural language processing through software solutions, sign language recognition technology is still evolving. Existing approaches to sign language recognition can be categorized into two main types: sensor-based and image-based solutions [19]. Image-based methods rely on cameras to capture hand movements and positions, which are then processed by software to interpret the signs. Although effective, these solutions often come with high costs, complexity, and limited flexibility for development and customization [26]. Moreover, their reliance on machine learning algorithms produces varying results depending on the techniques used, leaving room for further refinement [8].

On the other hand, Sensor-based solutions detect hand and finger movements using specialized sensors. However, these systems typically require users to wear gloves or other external hardware, which can be uncomfortable and impractical for everyday use [9], [16]. As a result, both image-based and sensor-based approaches face challenges that limit their real-world applicability.A promising alternative is the **Leap Motion Controller**, a compact device that connects to a computer and tracks hand movements without requiring any additional wearables. This innovative technology has gained attention in the field of sign language recognition due to its affordability, ease of use, and potential to overcome the limitations of traditional sensor-based and image-based systems [5].

This paper integrates Leap Motion and image-based recognition to capture the Arabic Sign Language (ArSL), where some signs rely not only on hand movements but also on the specific body parts the hand refers to. The Leap Motion Controller is utilized for its high precision in tracking hand motion, while a camera detects which body part the hand interacts with, such as the head, shoulder, or chest, since each reference point carries a different meaning in ArSL. For that reason, in this paper, we will investigate using multimodal. The goal is to develop an intuitive and efficient sign language recognition system that serves as a practical tool for the Deaf and Hard of Hearing (D/HoH) community.

Ultimately, this paper aims to develop Arabic sign language recognition technology, fostering accessibility and promoting social inclusion for the Deaf community across the Arab world.

This research paper aims to investigate the feasibility of a Multimodal Recognition System for Arabic Sign Language (ArSL) by evaluating the potential benefits of combining hand gesture recognition with contextual visual cues. To achieve this goal, several objectives have been outlined. Firstly, the system will utilize Leap Motion for precise 3D hand tracking and an RGB camera to capture body-referenced gestures, addressing observed limitations in single-sensor setups. Secondly, an AI model will be developed to classify and recognize ArSL signs, enabling an assessment of recognition accuracy, robustness, and system responsiveness under multimodal input. Finally, we compare the recognition accuracy performance of multimodal vs. single-sensor. Through this approach, the paper seeks to explore whether integrating multiple input modalities can explore the feasibility of accurate recognition performance in ArSL interpretation against a single model.

The paper addresses four key questions. Firstly, what are the key challenges in developing an Arabic sign language recognition system? It investigates the primary challenges involved in an Arabic sign language recognition system. By identifying these challenges, the research aims to establish a comprehensive understanding of the obstacles that need to be overcome. Secondly, what are the limitations of a single modal approach? This paper explores the limitations of single modality ,to understanding the capability of each modality for developing effective solutions. Lastly, how can to reduce or solve these limitations by using multimodal approaches? And how does the performance of a multimodal recognition system compare to a single modality in terms of accuracy in recognizing Arabic Sign Language (ArSL)? The research focuses on methods to reduce or resolve these challenges by thoroughly studying the reasons behind each modality. By conducting a detailed analysis, the paper aims to propose viable solutions that can to investigate the feasibility of a Multimodal Recognition System for Arabic Sign Language (ArSL) If we combine multiple input modalities such as 3D hand tracking and contextual visual cues then we can significantly improve the performance of Arabic Sign Language recognition compared to using a single modality.

## 2 LITERATURE REVIEW

Sign languages vary across regions, each having unique grammar, vocabulary, and cultural influences. For instance, Arabic Sign Language differs significantly from American Sign Language (ASL) and British Sign Language (BSL) in terms of gestures and contextual nuances. Multimodal approaches, such as [6], demonstrate the potential to integrate features across languages using tools like Leap Motion Controller (LMC) and RGB cameras. Transfer learning in these contexts has proven effective, as seen in the improved classification accuracy for ASL gestures using BSL-trained models. Arabic Sign Language (ArSL) recognition has garnered attention as a vital means of improving communication for the Deaf community in the Arab world. With the increasing need for assistive technologies, various methodologies have been explored, including sensor-based, image-based, and hybrid systems. Despite progress, there remains a gap in developing systems that are practically deployable and capable of addressing the unique challenges of ArSL.

### 2.1 Existing LMC-Based Systems in Sign Language Recognition

Among the sensor-based technologies, the **Leap Motion** has emerged as a promising tool due to its ability to track hand movements with high precision, without requiring gloves or additional wearable devices. Several studies have utilized the Leap Motion Controller (LMC) are summarized in Table 1, for recognizing static gestures in Sign Languages.

#### 2.1.1. LMC-Based Systems for Non-Arabic Sign Languages

Also, Researchers in [3] employed LMC to recognize static American Sign Language gestures, specifically targeting the alphabet and numbers. They utilized distance algorithms for the classification of the recorded gestures. Similarly, [4] explored the use of LMC for recognizing static gestures of American Sign Language but implemented recurrent neural networks (RNNs) as their classification method. This approach contrasts with the distance-based algorithms used by [3], highlighting a shift towards more sophisticated classification techniques. In another study, [17] extended the application of LMC by developing a system capable of recognizing both manual and finger-spelling gestures. A dataset of 2240 sign gestures consisting of 28 isolated manual signs and 28 finger-spelling words, This system incorporated a two-tier classification process: the first level utilized a Support Vector Machine (SVM) to differentiate between manual and finger-spelling gestures, while the second level employed an RNN for gesture recognition. While [17] have the largest number of signs that have gestures compared to other ArSL researches, but Overall accuracy was 63.57% in real-time recognition of sign gestures, which is a low accuracy. Furthermore, [9] also utilized LMC but focused on classifying 26 English alphabet gestures. They applied both k-nearest neighbor and SVM classification methods to achieve their recognition goals, demonstrating the versatility of classification approaches in the context of gesture recognition.

#### 2.1.2 LMC-Based Systems for Arabic Sign Language Recognition

In [19], a Leap Motion-based SL recognition system was introduced, which tracks hand movements for recognition of Arabic sign alphabets, gives 98% classification accuracy with the Nave Bayes classifier and more than 99% using the MLP.The researchers in, [15] developed a part-based hand gesture recognition system using LMC combined with a Support Vector Machine (SVM) classifier to identify static hand gestures corresponding to the Arabic letters "alif" - "yah" and digits 0-9. Similarly, [19] employed LMC for recognizing letters in ArSL but compared two different classifiers: the Naïve Bayes classifier and a Multilayer Perceptron. Their findings indicated that the neural network-based approach yielded superior results, but both research have limited dataset because just focus on letters and digits.

Moreover, [14] proposed a comprehensive model for both static and dynamic Arabic sign recognition using LMC. Their system covered a wide range of gestures, including 28 Arabic alphabet signs, digits 0-10, eight common signs used in dental contexts, 20 common nouns and verbs, and 10 two-handed signs. They evaluated the performance of three classifiers: SVM with a polynomial kernel, K-Nearest Neighbor (KNN), and Multilayer Perceptron. For static gestures, the classification methods achieved an accuracy of over 95%. For dynamic gestures, the suggested methods resulted in a good performance and accuracy of 98%. Additionally, a segmentation method based on motion speed achieved an accuracy of 95% in real-time continuous sign recognition. In another study, [11] utilized dual LMCs to recognize 100 different dynamic Arabic signs, employing both Bayesian approaches and Gaussian Mixture Models for classification. The model achieved an accuracy of 92%, but it has the limitation of using dual LMCs, where one LMC is positioned in front of the signer while the other is placed to the side, which makes this approach impractical. This setup can negatively impact the user experience.

Table 1: Summary of Existing LMC-Based Systems in Sign Language Recognition

| Research Reference Number | Language | Classification Method | Input Device | Word / Alphabet/Digits |
|---|---|---|---|---|
| 3 | American English | Algorithms | LMC | Digits / Alphabet |
| 4 | American English | RNN | LMC | Word / Digits / Alphabet / |
| 9 | American English | KNN, SVM | LMC | Alphabets |
| 11 | Arabic | NB ,GM | Dual-LMC | Words |
| 14 | Arabic | SVN , KNN ANN | LMC | Words / Alphabet /Digits |
| 15 | Arabic | SVM | LMC | Alphabet /Digits |
| 17 | American English | SVM,<br>BLSTM-NN | LMC | Words |
| 19 | Arabic | MLP ,Neural Networks ,<br>Nave Bayes Classifier | LMC | Alphabets |

## 2.2 Alternative Approaches to Sign Language Recognition

Sign Language (SL) recognition systems have been extensively studied; however, there remains a gap in developing systems that are practically deployable. Glove-based approaches have been explored in [22], alongside several image-based methods as proposed in [13,20,21,23,35]. Various approaches for sign language recognition have been explored as summarized in Table 2 , including sensor-based, image-based, and hybrid systems. Despite progress, there remains a gap in developing systems that are practically deployable and capable of addressing the unique challenges of ArSL.

### 2.2.1 Alternative Approaches for Non-Arabic Sign Language Recognition

Traditionally, sensor-based solutions involved the use of gloves equipped with sensors to detect hand movements and gestures. While effective, these systems have significant drawbacks, such as discomfort and impracticality for daily use. In[24] systematic review one of research explored sensor-based techniques, but they highlighted the limitations of glove-based systems, particularly for Arabic Sign Language (ArSL) recognition, such as detecting occlusion or the detection and segmentation of the hand and fingers. due to the difficulty in obtaining natural, fluid gestures when sensors are worn.

Similarly to research [19], Chai et al. [7] developed a sign language recognition and translation system utilizing depth and color images obtained from a Microsoft Kinect (MK) device. Their method aligns and matches the trajectory of each sign language gesture between probe and template images to determine the recognized result, achieving recognition rates of 83.51% and 96.32% on a dataset of 239 Chinese SL words using ranking approaches.

The research [6] presented in this study focuses on the recognition of British Sign Language (BSL) and American Sign Language (ASL) using a multimodal approach that integrates data from a Leap Motion Controller (LMC) and an RGB camera. The system is designed to classify a set of 18 common gestures from both sign languages, enabling effective communication for users who rely on sign language. The classification methods employed include deep learning techniques, specifically convolutional neural networks (CNNs), and the study highlights the advantages of a late fusion approach that combines features from both sensors.The performance of the model was rigorously evaluated, achieving an impressive accuracy of 94.44% for BSL classification. Additionally, the research explored transfer learning, demonstrating that models trained on BSL could significantly enhance the classification accuracy for ASL gestures, with the best model achieving 82.55% accuracy. The findings underscore the potential of multimodal systems in improving sign language recognition and suggest avenues for future research, including the expansion of gesture classes and further optimization of hyperparameters for enhanced performance.

The research[**32**] on gesture recognition for American Sign Language (ASL) leverages sEMG signals and machine learning to enhance communication technologies for Deaf and hard-of-hearing communities. By applying the CatBoost algorithm for gesture classification, the study achieved high accuracies of 99.99% for ASL-10 and 99.91% for ASL-24 datasets, demonstrating the robustness of this approach. Feature extraction involved approximately 450 time and frequency domain features from each sEMG channel, with Fast Fourier Transform coefficients playing a significant role. An ensemble feature selection process combined methods like ANOVA, Chi-square, Mutual Information, and ReliefF, optimized through a novel feature combiner to handle dimensionality. Data was collected from 20 subjects using a Myo armband across ASL digits and the manual alphabet, providing a reliable dataset. This study highlights sEMG's potential as a powerful alternative to computer vision in gesture recognition, advancing accessibility for ASL users.

The research [18] on Japanese Sign Language (JSL) recognition aims to improve communication for hearing-impaired individuals by developing a robust system capable of interpreting complex sign gestures. Using a hybrid model of Convolutional Neural Networks (CNN) and Bidirectional Long Short-Term Memory (BiLSTM) networks, the system captures spatial features through CNN and temporal dependencies through BiLSTM, effectively learning both static and dynamic signs. The study employs a multimodal data fusion method, combining keypoint hand joint angles from an RGB camera with 2-axis bending sensor data from gloves, enhancing recognition accuracy and resilience against background noise and occlusions. Tested on 32 JSL movements, the system improved recognition accuracy from 68.34% to 84.13%, showcasing its potential as a communication aid with applications in education, healthcare, and social integration.

### 2.2.2 Alternative Approaches for Arabic Sign Language Recognition

In contrast, some researchers have applied image processing techniques to Arabic sign language recognition[1,2,13,20,21]. For instance, [1] proposed a method for processing images of Arabic alphabet gestures to extract features that were subsequently classified using SVM. Likewise, [2] utilized image processing to extract features for recognizing ArSL but focused on gestures representing words rather than individual letters. Their approach involved Hidden Markov Models (HMM) for classification. The average sentence recognition rate was 75% and the word recognition rate was 94%.

Dynamic gesture recognition in ArSL has also been an area of interest. [13] developed a system for recognizing dynamic Arabic sign language using Microsoft Kinect. This system employed two machine learning algorithms—Decision Tree and Bayesian Network—and applied AdaBoosting to enhance recognition performance.Also, Sensor-based systems for ArSL that use gloves have been explored in [22] but can be uncomfortable and non-portable, making them impractical for daily use.

Table 2: Summary of Alternative Approaches to Sign Language Recognition

| Research Reference Number | Language | Classification Method | Input Device | Word / Alphabet/Digits |
|---|---|---|---|---|
| 1 | Arabic | SVM | Camera | Alphabet |
| 2 | Arabic | HMM | Camera | Words |
| 6 | British | CNN-MLP-LF<br>Transfer Learning | LMC and Camera | Words |
| 7 | Chines | 3D trajectory matching algorithm | knicet | Words |
| 13 | Arabic | DT NB | knicet | Words |
| 18 | Japanese | CNN<br>BiLSTM | Monocular RGB Camera<br>And Bending sensor Data Gloves | Words / Alphabet |
| 21 | Arabic | HM | Camera | Words |
| 22 | Arabic | SVM | Glove | Words |
| 32 | American English | CatBoost algorithm | EMG sensors | Alphabet/Digits |
| 35 | Chines | SVM | Camera | Alphabet |

### 2.3 Challenges of Arabic Sign Language Recognition

While significant progress has been made in sign language recognition (SLR), existing methods often fall short in practical deployment and addressing the unique challenges of Arabic Sign Language (ArSL). There are some limitations in prevalent approaches: First, limited handling of dynamic gestures: Sensor-based systems that use gloves can be uncomfortable and non-portable, making them impractical for daily use [24]. Also, dual-LMC was also impractical for daily use [11]. Additionally, they may struggle to capture natural, fluid gestures due to limitations in sensor fidelity [24]. Camera-based approaches might miss subtle hand and finger movements crucial for accurate ArSL recognition. Second, Performance Bottlenecks: Existing SLR systems, particularly those focused on ArSL, often achieve accuracy plateaus using single-sensor modalities (camera or Leap Motion) [15,19, 14] while research [14] achieved good performance, but the sign language dataset included small scope and lacked variety, as it mainly focused on hand gestures. Some signs in real-world applications require other parts of the body, but the Leap Motion Controller (LMC) is unable to capture those parts. Also, some research just focused on letters and digits recognition [15,19]. These limitations can be attributed to the inherent complexities of ArSL, characterized by rich variations in hand shapes, movements, and cultural nuances. Third, Data Scarcity: Training robust deep learning models for SLR necessitates vast amounts of labeled data. Limited availability of high-quality ArSL datasets can hinder the generalization capabilities of these models [24]. One of the most significant challenges in Arabic Sign Language Recognition (ArSLR) research is the scarcity of large, annotated datasets. This is a recurring theme in ArSL research, often mentioned as a primary obstacle to achieving high recognition accuracy. Existing datasets [2] are often limited in size, containing a restricted number of signs, signers, and variations in signing styles, which hinders the development of robust and generalizable recognition models. This data scarcity can lead to overfitting, where models perform well on training data but poorly on unseen data, limiting performance in real-world scenarios. Several strategies can mitigate these issues. Data augmentation techniques, such as image transformations (rotation, flipping, cropping) and temporal modifications (speed variation, frame skipping), can artificially increase dataset size and improve model robustness [30]. This is a general technique applicable to any image/video-based recognition task where data is limited. Transfer learning offers another promising avenue. Leveraging pre-trained models on large-scale image or video datasets provides a strong initial representation that can be fine-tuned for ArSLR, especially beneficial with limited labeled data [27]. This approach has been successfully applied in various low-resource language processing tasks. Furthermore, the concept of cross-lingual transfer learning, while less explored in ArSL specifically, could be valuable. The idea is that knowledge learned from other, larger sign language datasets could be transferred to ArSL, as sign languages share underlying visual and temporal patterns. This approach has been successfully demonstrated in other sign language recognition tasks, such as transferring knowledge from American Sign Language to British Sign Language [6]. Finally, active data collection initiatives are crucial. Crowdsourcing through mobile apps and online platforms can engage the ArSL community and increase data volume and diversity. Collaborations between researchers and organizations can also facilitate larger-scale data collection and data sharing within the ArSLR community. Addressing these dataset limitations through these combined approaches is crucial for advancing ArSLR and developing more accurate and reliable recognition systems. By acknowledging these shortcomings, we can pave the way for developing more practical and robust SLR systems, particularly for resource-constrained environments and languages like ArSL.

## 3 METHODOLOGY

After exploring existing technology, as summarized in Table 3 for sign language recognition, the proposed solution leverages both Leap Motion and an RGB camera to address the limitations of single-sensor approaches that most researchers followed, as shown in Table 3 to recognize Arabic Sign Language(ArSL) . The proposed multimodal system leverages Leap Motion for detailed 3D hand tracking and RGB cameras for capturing contextual visual information. This integration is designed to address the limitations of single-sensor systems by combining spatial precision with contextual awareness, enhancing the recognition of both static and dynamic gestures in ArSL. Indeed, Leap Motion provides detailed 3D hand tracking data, capturing the precise movements and positions of the hands, which is crucial for recognizing sign language gestures. It is also a portable device, making it easy to carry anywhere in real life, unlike other sensors. On the other hand, RGB cameras capture visual information, including the context and appearance of the signer, which can help in understanding the signs better. By combining these two modalities, the system can leverage the strengths of each to improve overall recognition accuracy. Furthermore, the proposed framework combines features using CNNs and Transfer learning to classify gestures effectively. Through this approach, the research aims to investigate whether integrating multiple input modalities can explore the feasibility of accurate recognition performance in ArSL interpretation compared to a single model by evaluating the accuracy of multimodal models against single-modal models.This proposed solution, inspired by [6] and [18], integrates spatial and temporal data for robust recognition of ArSL gestures. The dual-sensor setup enhances accuracy and broadens the range of recognizable gestures, particularly for dynamic and contextually rich sign languages like ArSL.

Table 3: Summary of All Approaches to Sign Language Recognition

| **Research Reference Number** | **Language** | **Classification Method** | **Input Device** | **Word / Alphabet/Digits** |
|---|---|---|---|---|
| 1 | Arabic | SVM | Camera | Alphabet |
| 2 | Arabic | HMM | Camera | Words |
| 3 | American English | Algorithms | LMC | Digits / Alphabet |
| 4 | American English | RNN | LMC | Word / Digits / Alphabet / |
| 6 | British | CNN-MLP-LF<br>Transfer Learning | LMC and Camera | Words |
| 7 | Chinese | 3D trajectory matching algorithm | knicet | Words |
| 9 | American English | KNN, SVM | LMC | Alphabets |
| 11 | Arabic | NB ,GM | Dual-LMC | Words |
| 13 | Arabic | DT NB | knicet | Words |
| 14 | Arabic | SVN , KNN ANN | LMC | Words / Alphabet /Digits |
| 15 | Arabic | SVM | LMC | Alphabet /Digits |
| 17 | American English | SVM,<br>BLSTM-NN | LMC | Words |
| 18 | Japanese | CNN<br>BiLSTM | Monocular RGB Camera<br>And Bending sensor Data<br>Gloves | Words / Alphabet |
| 19 | Arabic | MLP neural networks ,<br>Nave Bayes classifier | LMC | Alphabets |
| 21 | Arabic | HM | Camera | Words |
| 22 | Arabic | SVM | Glove | Words |
| 23 | Arabic | HMD | Camera | Words |
| 32 | American English | CatBoost algorithm | EMG sensors | Alphabet/Digits |
| 35 | Chinese | SVMs | Camera | Alphabet |

This study adopts a quantitative, experimental research design following a deductive (top-down) approach where we conduct the hypothesis after that, we will test it through evaluation of Arabic Sign Language recognition multimodal models using the F1-score, confusion matrix, and classification report is critical to ensure the robustness and reliability of the proposed system. The F1-score, which balances precision and recall, is particularly effective in scenarios with imbalanced datasets, as it highlights the trade-off between correctly identified gestures and the false positives or false negatives. This is essential for Arabic Sign Language recognition, where the misclassification of gestures can lead to significant misinterpretation. The confusion matrix provides a detailed analysis of model performance by showing the distribution of true positives, false positives, true negatives, and false negatives, offering insights into specific gesture pairs that the model struggles to differentiate. Lastly, the classification report provides a comprehensive summary of precision, recall, and F1-score for each class, ensuring that the model's performance is thoroughly assessed across all gestures, both static and dynamic. These metrics collectively provide a holistic evaluation of the system, ensuring its effectiveness and reliability in accurately recognizing Arabic Sign Language gestures in real-world applications. The final phase of the study involves a comparative analysis between multimodal and single-modal models, by compare accuracy of multimodal models against single-modal models.

### 3.1 Data Acquisition

First a custom Saudi Sign Language dataset was developed through simultaneous data collection using a Leap Motion sensor and a standard webcam. The dataset includes 18 distinct sign language gestures, based on references from [28], which include both one-handed and two-handed signs. These gestures were selected based on their frequency of use in daily communication. With each gesture performed ten times, during each repetition, 73 image frames were captured along with 73 synchronized motion data entries from the Leap Motion sensor, ensuring alignment between visual and motion data. The extracted Leap Motion features are summarized in Table 5, and the gesture classes are listed in Table 4.

Table 4: Dataset of Sign Language Recognition

| Word Number | Arabic Meaning | English Meaning |
|---|---|---|
| 1 | السلام عليكم | Alsalam Ealaykum |
| 2 | وعليكم السلام | Wa Ealaykum Alsalam |
| 3 | مع السلامة | Good bye |
| 4 | أهلا وسهلا | Hello |
| 5 | اسم | Name |
| 6 | أنا آسف | I am Sorry |
| 7 | لو سمحت | Please |
| 8 | شكراً | Thanks |
| 9 | وقت | Time |
| 10 | سئ | Bad |
| 11 | حسناً | ok |
| 12 | باص | Bus |
| 13 | سيارة | Car |
| 14 | سيارة أجرة | Taxi |
| 15 | طيارة | Plane |
| 16 | قطار | Train |
| 17 | طعام | Food |
| 18 | شراب | Drink |

For each gesture, images were recorded at a frequency of one every 0.2 seconds using a laptop webcam, while Leap Motion data was collected simultaneously from a device positioned above the camera, facing the participant. The setup allowed participants to experience direct communication through a simulated interaction with another person. Each 0.2-second interval yielded a data sample consisting of both an image and its corresponding motion data, which were stored as numerical vectors for subsequent classification.

The dataset captures comprehensive hand-tracking features using a Leap Motion Controller (LMC), structured to support the development of an Arabic Sign Language (ArSL) recognition system. These features are extracted from each frame and include the following components as shown in Table 5 and depicted in Figure 1.

Table 5: Feature Extraction from Leap Motion

| Feature Category | Details |
|---|---|
| **Arm Features** | - Start and end positions (x, y, z) |
| | - 3D angle between start and end points |
| **Palm Features** | - Position and velocity vectors<br>- Normal vector orientation<br>- Pitch, roll, and yaw angles<br>- Palm angle relative to the normal vector |
| **Finger Features** | - Tracked for all five fingers<br>- For each bone (metacarpal, proximal, intermediate, distal):<br>● Start and end joint positions.<br>● Bone width.<br>● 3D angle of the bone. |
| **Hand Presence Flags** | - Indicates whether left and/or right hands are detected |

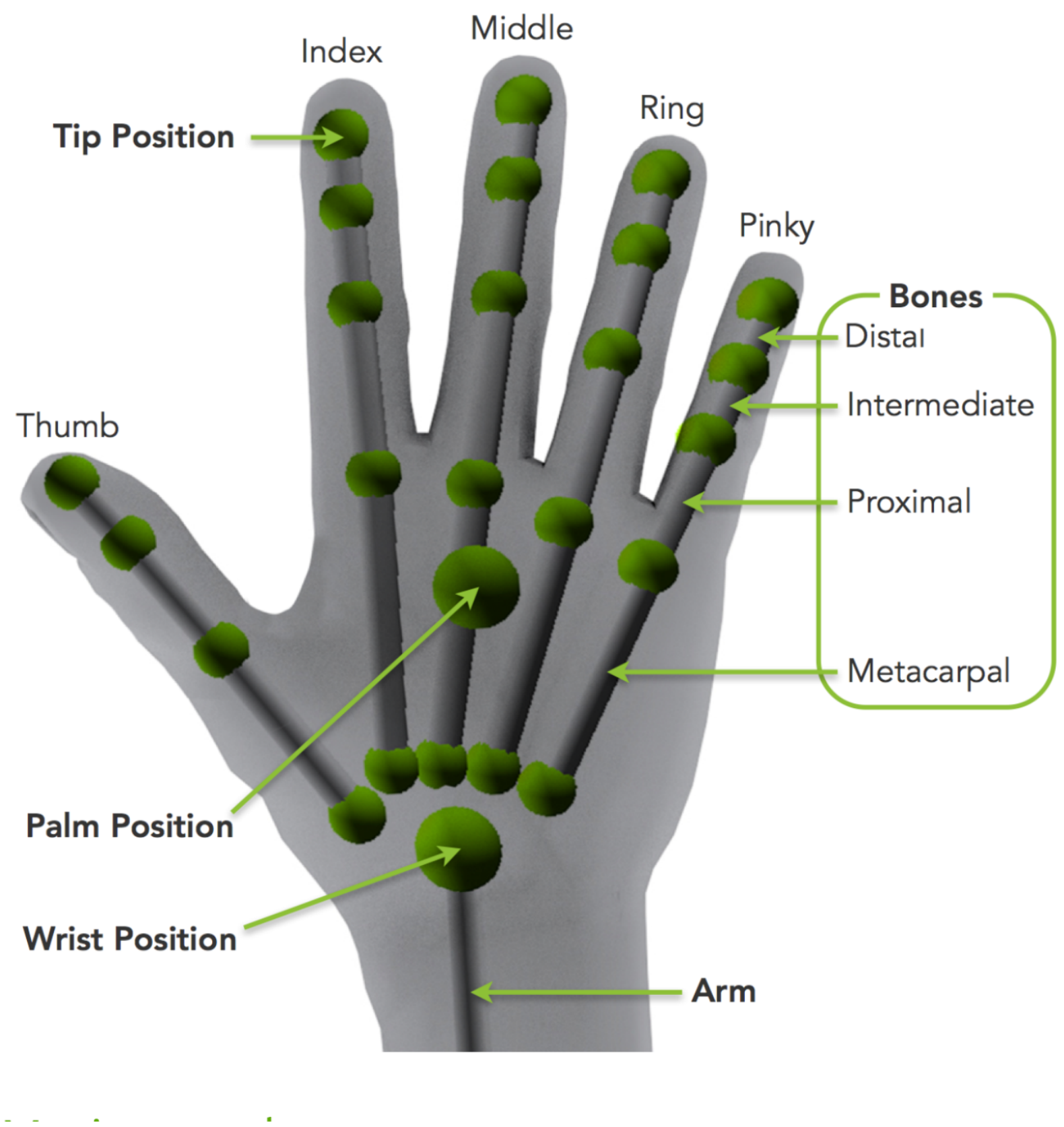

Figure 1: Features of a Hand

- **Arm Features**: Include the 3D coordinates (x, y, z) of the start and end positions of the arm, along with the 3D angle formed between them.

- **Palm Features**: Encompass position and velocity vectors, the orientation of the palm's normal vector, and the computed pitch, roll, and yaw angles. Additionally, the angle between the palm and its normal vector is included to capture wrist and hand orientation.

- **Finger Features**: Tracked for all five fingers, with data collected for each of the four bones per finger (metacarpal, proximal, intermediate, and distal). For each bone, the dataset records:

    - Start and end joint positions (x, y, z)

    - Bone width

    - 3D angle of the bone

- **Hand Presence Flags**: Indicate whether the left and/or right hand is present in each frame.

The leap motion features data is organized by sign name and stored in CSV format for ease of processing in the Arabic Sign Language dataset development.

To compute the 3D angles (θ) between vectors such as between bones or arm segments the following formula is used:

$$\theta = arccos(a \cdot b \,/\, | a |\cdot| b |)$$

Where:

- $|\vec{a}|$ *and* $|\vec{b}|$ are the magnitudes of vectors **a** and **b**, respectively, calculated as:

$$|\vec{a}| = \sqrt{a_x^2 + a_y^2 + a_z^2}\ ,\ |\vec{b}| = \sqrt{b_x^2 + b_y^2 + b_z^2}$$

  Here, each vector is defined by the start and end points of a bone or segment in 3D space, as provided by the Leap Motion Controller.

## 3.2 Data Preprocessing

To ensure the quality and consistency of the input data, several preprocessing steps were applied to the data.

### 3.2.1 Leap Motion Features Preprocessing

There are some preprocessing steps applied to the features collected from the Leap Motion Controller. First, any missing values (NaNs) were replaced with zeros to maintain the integrity of the dataset. Following this, all numerical features were normalized to a range between 0 and 1 using Min-Max Scaling. The scaling formula used is as follows:

$$\text{Featscaled} = \frac{\text{Feat} - \text{Feat}_{\text{min}}}{\text{Feat}_{\text{max}} - \text{Feat}_{\text{min}}}$$

The normalization process was necessary because raw, unscaled feature values led to poor model performance, with classification accuracy dropping to approximately 5%, indicating a failure to generalize effectively.

### 3.2.1 Image Preprocessing

For the visual data, each image was resized to 224×224 pixels to match the input size requirements of standard deep learning models. Subsequently, pixel values were scaled to the [0, 1] range by dividing each pixel value by 255.

## 3.3 Model Architecture and Training

The proposed Fusion Model for Arabic Sign Language (ArSL) recognition has been developed using a multi-model architecture that integrates both motion and image-based data sources, as shown in Figure 2. The training process employed a supervised learning approach using a curated dataset of 18 ArSL signs, collected via Leap Motion and image inputs. The Fusion Model was trained using the cross-entropy loss function and optimized with the RMSprop algorithm. Training was conducted over multiple epochs, with a validation split employed to monitor model performance and prevent overfitting. Transfer learning was leveraged through the VGG16 backbone, enabling efficient feature extraction and reducing training time.

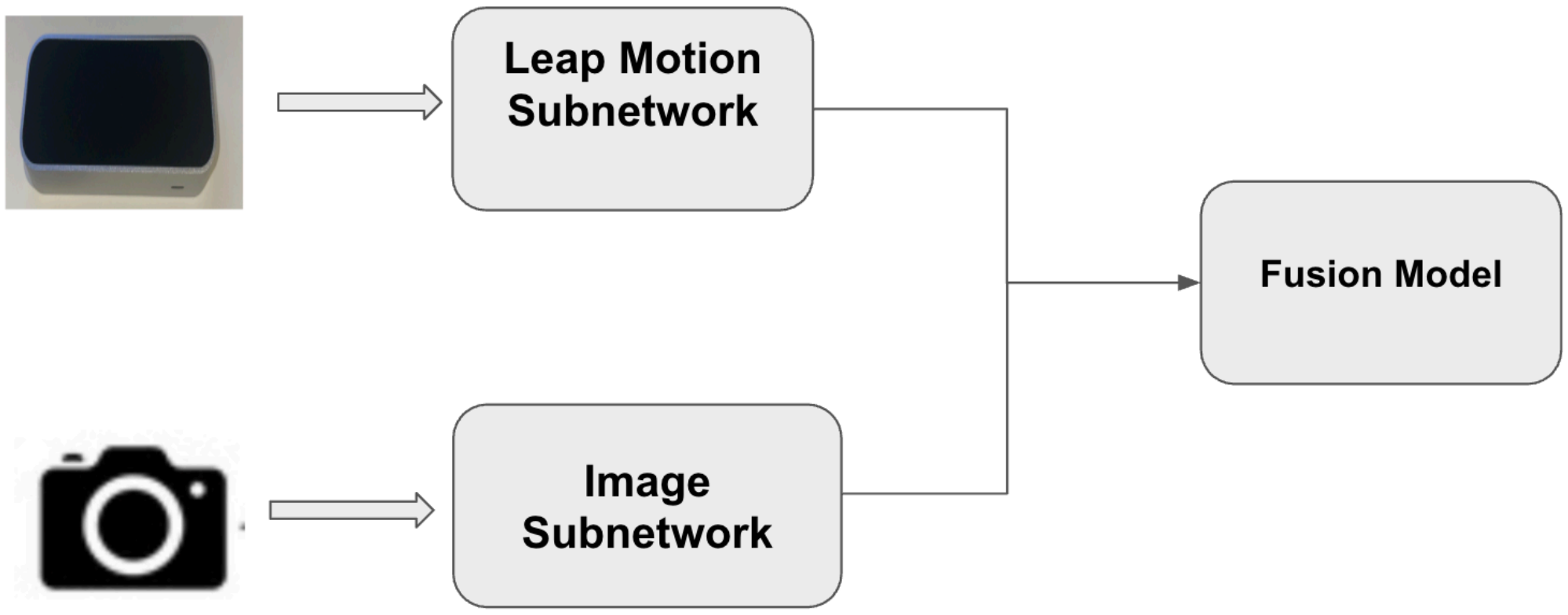

*Figure 2: Model Architecture*

### 3.3.1 Architecture Overview:

1. **Leap Motion Subnetwork:** This subnetwork is designed as a custom dense neural network that incorporates dropout layers and L2 regularization to prevent overfitting. It processes hand motion data captured from the Leap Motion Controller. It consists of:
   - A 512-unit dense layer with ReLU activation and L2 regularization.
   - A dropout layer with a rate of 0.2.
   - A 256-unit dense layer with similar regularization and dropout.
   - A final 128-unit dense layer followed by flattening.

2. **Image Subnetwork:** The image pathway utilizes a pre-trained VGG16 [31] model initialized with ImageNet weights. This model is fine-tuned by appending custom fully connected layers and applying data augmentation techniques such as random rotation, zoom, and contrast adjustments to enhance generalization.
3. **Fusion Model:** Features extracted from both subnetworks are concatenated and passed through additional fully connected layers. The final classification is achieved using a softmax activation function, producing probabilities for each sign class. It consists of:

   - A dense layer with 256 neurons and ReLU activation, followed by a dropout layer.
   - A subsequent dense layer with 128 neurons, followed by another dropout layer.
   - A final dense layer with classes of neurons and softmax activation for multi-class classification.

## 4 RESULT

### 4.1 Exploratory Data Analysis (EDA)

As part of the initial exploratory data analysis for the classification task, the dataset comprises a total of 18 distinct sign classes, each representing a unique gesture or label. For each class, there are exactly 10 samples as displayed in Figure 3, resulting in a total of 90 samples in the entire dataset. The data is split into three subsets: 70% for training, 15% for testing, and 15% for validation as displayed in Figure 4. This corresponds to 7 samples per class for training, 1 or 2 samples for testing, and 1or 2 samples for validation. While the dataset is relatively small and balanced across classes, the limited number of samples per class highlights the need for careful model regularization and potentially data augmentation techniques to enhance generalization and avoid overfitting during training.

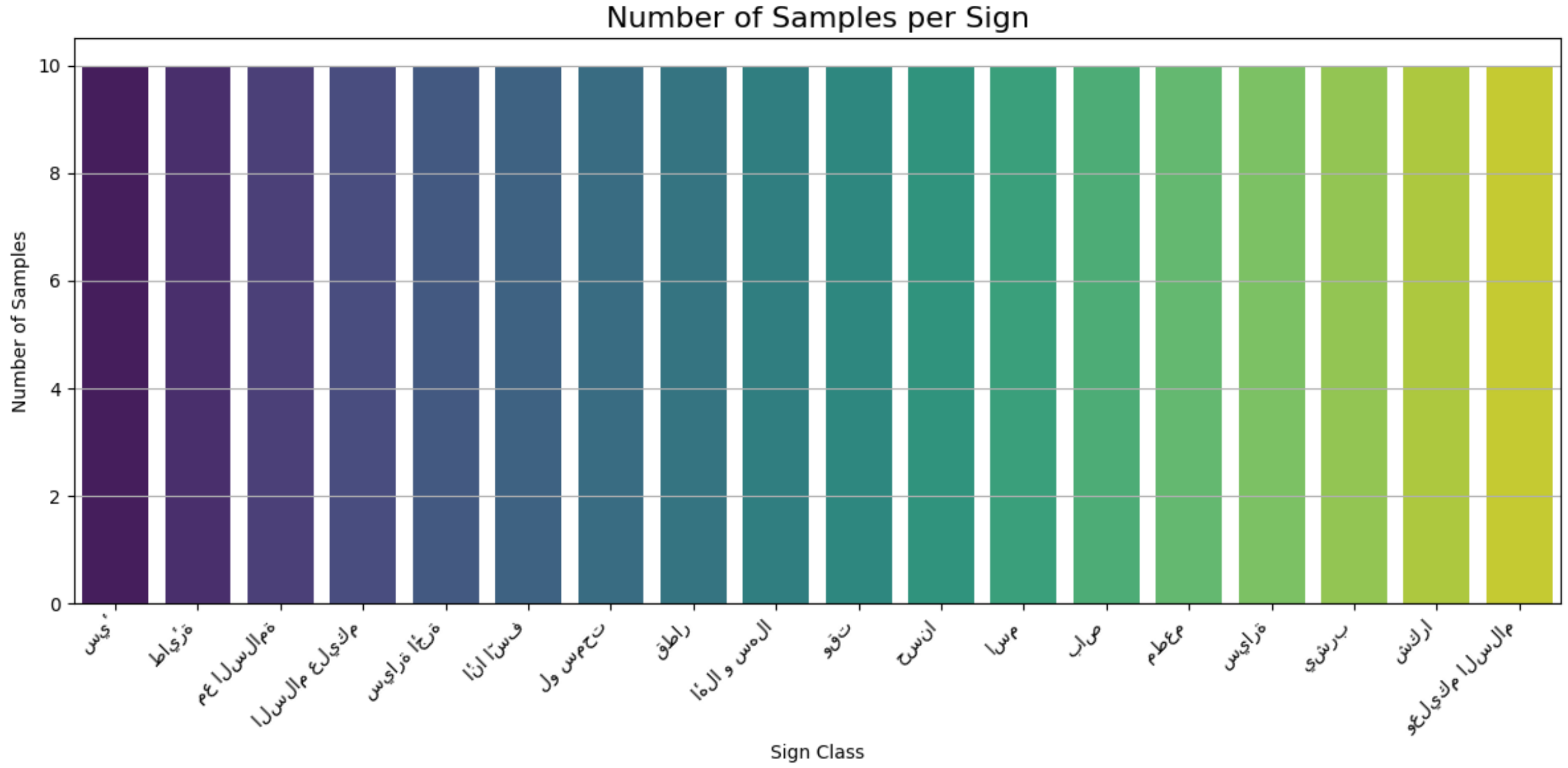

Figure 3: Number of Samples per Sign

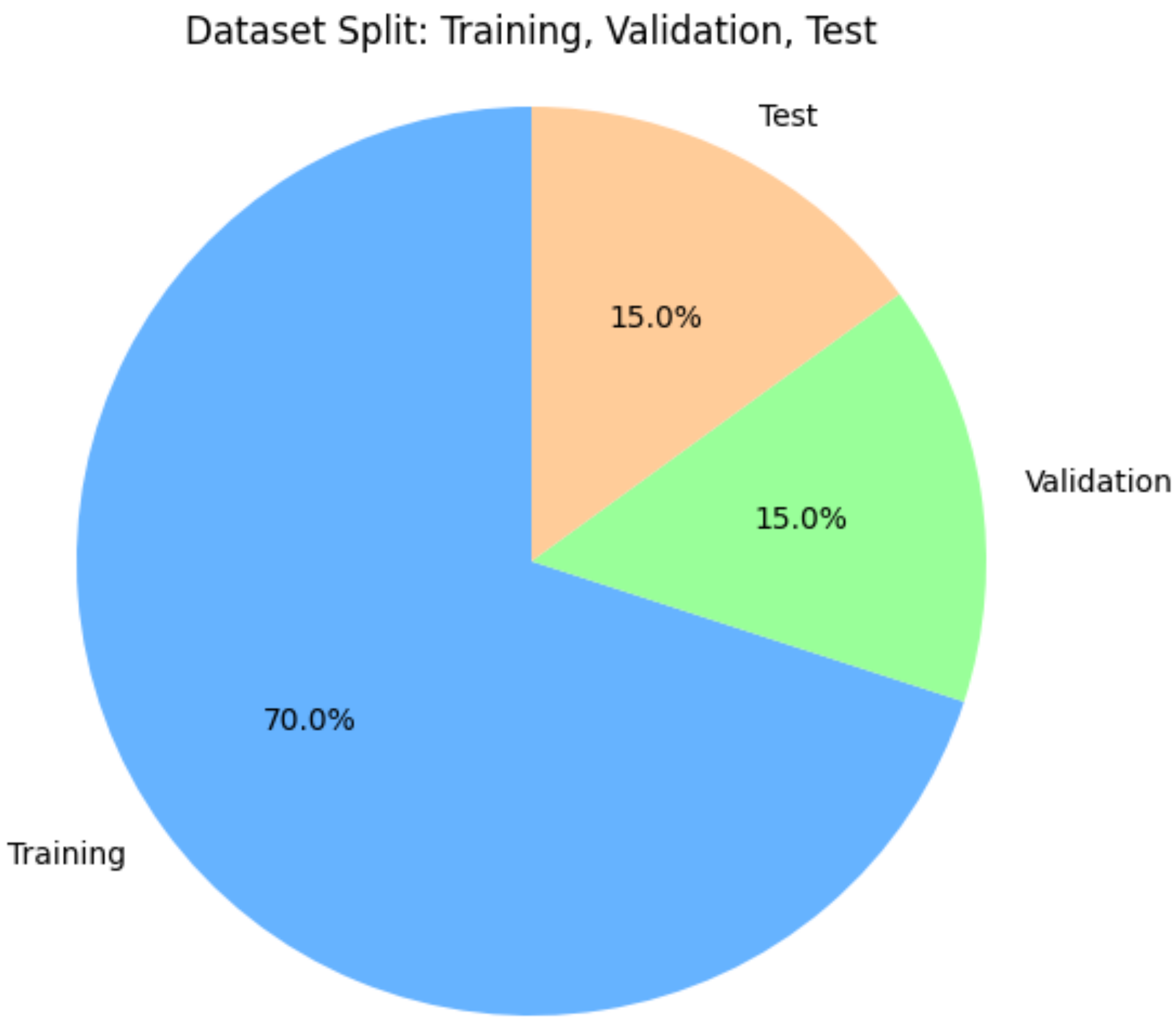

Figure 4: Dataset Split

### 4.2 Training Metrics and Performance

The complete dataset was split into three subsets: 70% for training, 15% for validation, and 15% for testing, to ensure robust model development and evaluation. Figure 5 (left) depicts the evolution of training and validation accuracy over 75 epochs, while Figure 5 (right) shows the corresponding loss curves.

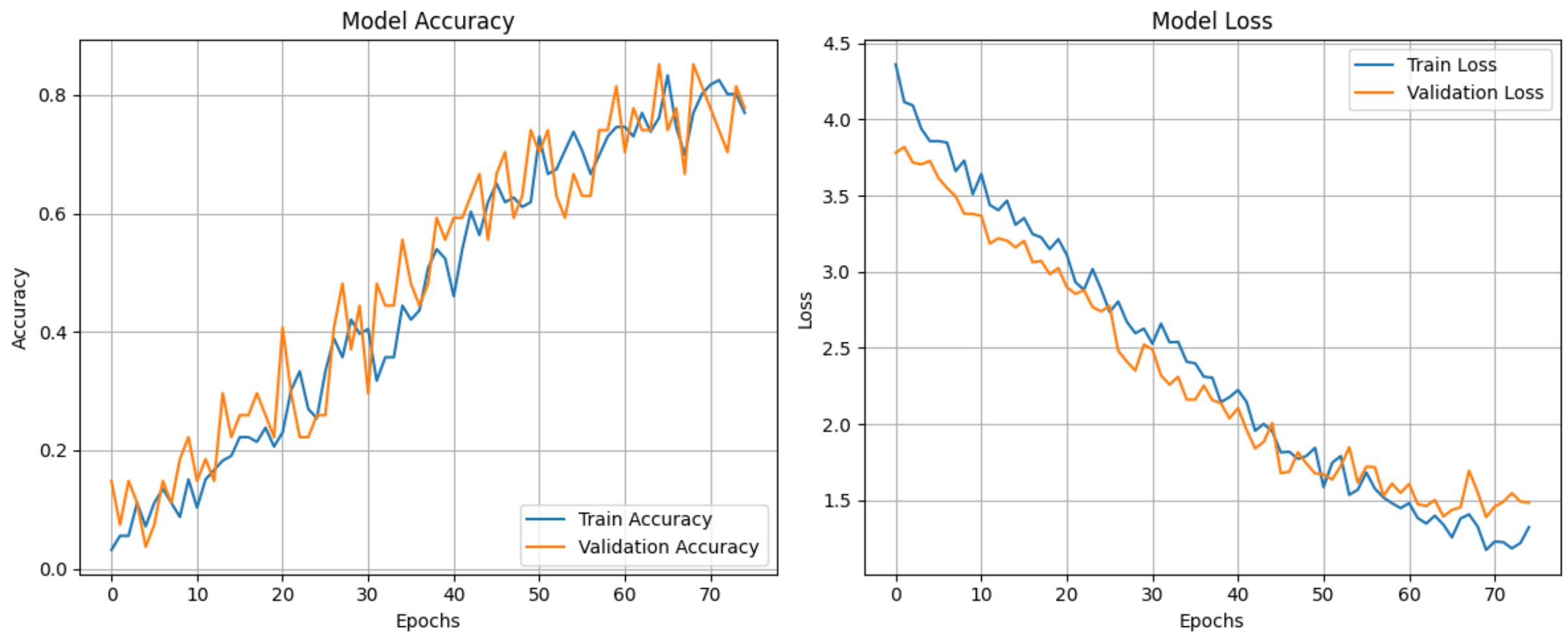

Figure 5: Accuracy and Loss of Model

The model demonstrates rapid loss reduction during first 10-15 epochs as it decreases training set loss from 4.4 to 3.5 and validation set loss from 3.8 to 3.2 while simultaneously increasing accuracy (train:5% - 15%, val: 10% - 22%) . The network demonstrates rapid decrease in loss during the initial stage because it quickly identifies basic patterns in data. The model demonstrates effective abstract feature extraction during mid-phase (epochs 15 to 40) because traing accuracy increases from 20% to 60%, and validation accuracy rises from 25% to 58% while validation and training losses remain closely aligned. The model achieves training accuracy of 0.80 and validation accuracy of 0.78 by epoch 75 during the late-stage convergence phase. Training loss stabilizes at 1.3 and validation loss stabilizes at 1.5 while validation metrics show minor fluctuation due to stochastic factors like batch sampling variance or learning-rate adjustments rather than overfitting.

The narrow gap between training and validation curves throughout especially in loss indicates the model generalizes well. There is no sustained divergence (i.e., the validation loss does not rise while training loss falls), which suggests capacity is well-matched to the dataset. Overall, the model demonstrates strong learning dynamics: rapid initial convergence, sustained mid-training gains, and stable late-stage performance.

### 4.3 Classification Results

The confusion matrix below in Figure 6 summarizes our model's performance on the held-out test set across 18 target categories. Each row corresponds to the true class, and each column to the predicted class; perfect classification appears on the main diagonal.

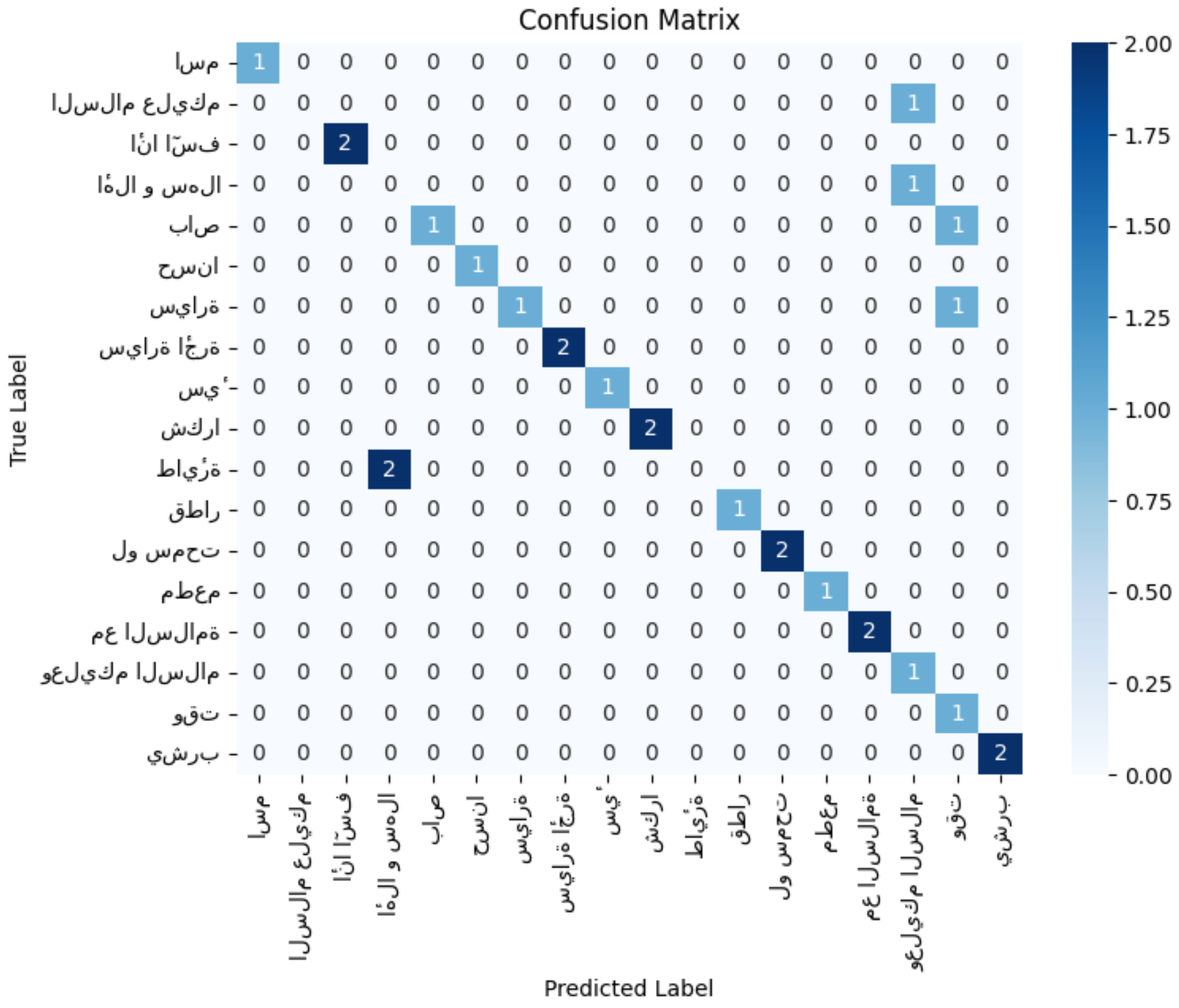

Figure 6: Confusion Matrix

The majority of test samples appear on the diagonal (either a "1" or "2" in each diagonal cell), indicating that most instances are correctly recognized. The model achieves correct classification for approximately 72 % of all samples are classified correctly, demonstrating strong overall accuracy on unseen data. In per-class performance, the model demonstrates good classification performance for several categories, as indicated by a diagonal count of “2” in the confusion matrix, reflecting perfect accuracy across both test instances per class; and some categories just have one sample and also appear on diagonal of the confusion matrix, reflecting perfect accuracy of test instances per class; this suggests that the model has effectively internalized the distinguishing characteristics of these classes. However, some samples not appearing on the diagonal refer to misclassification.

The bar chart shown in Figure 7 summarizes the model’s overall test-set performance across four key metrics:

- **Accuracy (78 %):**
  Accuracy confirms that nearly eight out of ten test samples are correctly classified.

- **Precision (80 %) vs. Recall (78 %):**
  Precision measures how many of the positively predicted labels are actually correct. At 0.80, the model makes a few false-positive mistakes when it predicts a class it is usually right. Recall measures how many of the true instances are successfully retrieved. At 0.78, the model detects most of the true class examples but still misses about 22 % of them. The slightly higher precision compared to recall indicates a conservative bias: the classifier errs more on the side of omission (false negatives) than over-assigning classes.

- **F1-Score (76 %):**
  The F1-Score harmonizes precision and recall into a single measure. A value of 0.76 reflects a balanced trade-off, confirming that neither false positives nor false negatives dominate the error profile.

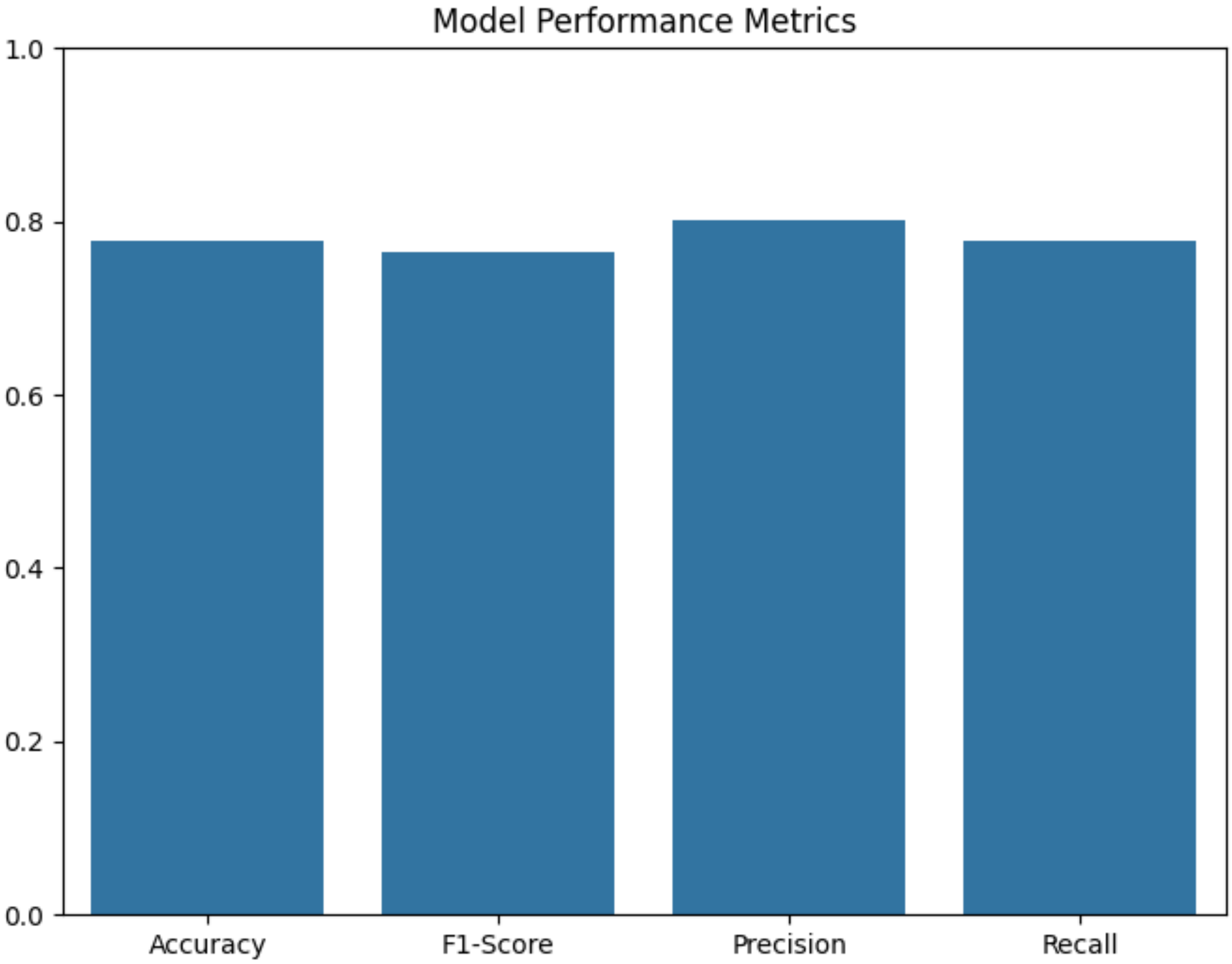

Figure 7: Performance Metrics of Model

As displayed in Table 6 the classification report provides a comprehensive evaluation of the model's performance across 18 distinct Arabic utterance classes. The overall accuracy achieved is 78%, indicating that a majority of the predictions align with the ground truth labels. High precision, recall, and F1-scores of 1.00 are observed for several classes such as "سيارة أجرة", "حسنا", "أنا آسف" and "شكرا", demonstrating the model’s robustness in recognizing these frequently or distinctly represented utterances. However, the model fails to identify certain classes, such as "أهلا و سهلا", "السلام عليكم" and "طائرة", with scores of zero across all evaluation metrics. This suggests either insufficient representation of these classes in the training data or model difficulty in distinguishing these semantically similar utterances. Because of the sign "طائرة", I believe the model may need more data to properly learn and distinguish it. As I’m not a specialist in sign language translation in typical scenarios, data collection is usually handled by professional sign language interpreters who are experts in this area. As for "أهلا وسهلا" and "السلام عليكم", their signs are quite similar, which might be contributing to the confusion. Furthermore, classes like "وعليكم السلام" and "وقت" exhibit a disparity between precision and recall, with recall being high but precision relatively low, indicating a tendency toward false positives for these categories. The macro-averaged F1-score of 0.74 highlights moderate overall performance with potential imbalance in class-wise predictive capabilities. Previous studies have demonstrated that Leap Motion technology often yields higher accuracy in sign recognition tasks compared to traditional image-based methods, particularly for signs that rely solely on hand shape and motion. However, one potential limitation that has not yet been fully addressed in existing research is the challenge of distinguishing signs that share identical hand configurations but differ in meaning based on the hand's spatial reference to specific body parts. For instance, some signs in Arabic Sign Language (ARSL) use the same hand gesture but convey different meanings depending on whether the hand points to the head, chest, or other regions. Since Leap Motion does not capture visual context beyond the hands, this scenario remains inadequately supported by single-modality systems. To overcome this, our proposed multimodal ARSL recognition system combines the high-accuracy 3D hand tracking capabilities of Leap Motion validated in prior research with visual data from a camera to interpret contextual body references. This integrated approach enables the system to differentiate between gestures that appear identical in hand shape but diverge semantically based on body positioning. Our preliminary evaluation achieved an overall accuracy of 78%, underscoring the potential of multimodal fusion for more robust and comprehensive ARSL recognition. These findings offer promising evidence of the viability of this approach while also indicating the need for further dataset expansion and refinement of the visual recognition pipeline to better capture context-dependent signs.

Table 6: Classification Report

| **Label** | **Precision** | **Recall** | **F1-Score** | **Support** |
|---|---|---|---|---|
| اسم | 1.0 | 1.0 | 1.0 | 1 |
| السلام عليكم | 0.0 | 0.0 | 0.0 | 1 |
| أنا آسف | 1.0 | 1.0 | 1.0 | 2 |
| أهلا و سهلا | 0.0 | 0.0 | 0.0 | 1 |
| باص | 1.0 | 0.5 | 0.67 | 2 |
| حسنا | 1.0 | 1.0 | 1.0 | 1 |
| سيارة | 1.0 | 0.5 | 0.67 | 2 |
| سيارة أجرة | 1.0 | 1.0 | 1.0 | 2 |
| سيئ | 1.0 | 1.0 | 1.0 | 1 |
| شكرا | 1.0 | 1.0 | 1.0 | 2 |
| طائرة | 0.0 | 0.0 | 0.0 | 2 |
| قطار | 1.0 | 1.0 | 1.0 | 1 |
| لو سمحت | 1.0 | 1.0 | 1.0 | 2 |
| مطعم | 1.0 | 1.0 | 1.0 | 1 |
| مع السلامة | 1.0 | 1.0 | 1.0 | 2 |
| وعليكم السلام | 0.33 | 1.0 | 0.5 | 1 |
| وقت | 0.33 | 1.0 | 0.5 | 1 |
| يشرب | 1.0 | 1.0 | 1.0 | 2 |
| **Accuracy** | | | 0.78 | 27 |
| **Macro Avg** | 0.76 | 0.78 | 0.74 | 27 |
| **Weighted Avg** | 0.8 | 0.78 | 0.77 | 27 |

## 5 DISCUSSION

This study investigates the potential of sensor fusion for gesture recognition and highlights the practical considerations involved in real-world deployment. The proposed multimodal system achieved a classification accuracy of 78% across 18 Arabic Sign Language (ArSL) word-level gestures. To assess its significance, these results can be compared to other sign language recognition systems, as shown in Table 7, which used different modalities, gesture sets, and classification approaches.

Table 7: Benchmarking Prior Research Against Our Method

| ***Study Reference*** | **Language** | **Modality** | **Sign Set** | **Accuracy** | **Comparison** |
|---|---|---|---|---|---|
| **1** | Arabic | RGB Camera | Alphabet | 63.5% | Lower, while task was simpler(letter classification). |
| **2** | Arabic SL | RGB Camera only | Words | 75% (sentences), 94% (words) | Comparable;some aspects better, but multimodal not used. |
| **3** | ASL | Leap Motion | Digits, Alphabet | Digits,100% Alphabet 94% | High, but task was simpler (letter classification, Number classification). |
| **4** | ASL | Leap Motion | Digits,<br><br>Alphabet | 96% | Higher accuracy, but smaller dataset and more controlled gestures. |
| **6** | ASL/BSL | Leap Motion , RGB Camera | 18 Words | BSL: 94.4%, ASL: 82.5% | Higher, but used Late Fusion and larger datasets. |
| **7** | Chinese SL | Kinect | Words | 96.32% | Higher, but Kinect is less portable because of its bulk, external power, and more complex setup.<br><br>. |
| **9** | ASL | Leap motion | Alphabet | 79.83% | Lower, while task was simpler (letter classification). |
| **11** | Arabic SL | Dual Leap Motion (2 LMCs) | 100 Words | 92% | Higher, but impractical setup with two sensors. |
| **13** | Arabic SL | Kinect | Words | 93.7% | Higher, but Kinect is less portable because of its bulk, external power, and more complex setup. |
| **14** | Arabic SL | Leap Motion | Words , Digits | >95% (static), 98% (dynamic) | Higher accuracy, but smaller dataset and more controlled gestures. |
| **15** | Arabic SL | Leap Motion | 28 Words,<br><br>Digits,<br><br>Alphabet | 91% | Higher accuracy, but smaller dataset more controlled gestures, most tasks was simpler (letter classification). |
| **17** | ASL | Leap Motion | 28 Words | 63.6% | Lower; our system outperforms it despite a similar sensor and more complex vocabulary. |
| **18** | Japanese SL | Gloves and RGB Camera | 27 Words Static Movements, 5 Dynamic Movements | 84.13% | Higher accuracy, but Data gloves are less portable and require the user to wear them, which can negatively impact the overall user experience |
| **19** | Arabic SL | Leap Motion | Alphabet | 98%–99% | Higher, but task was simpler (letter classification). |
| **21** | Arabic SL | RGB Camera | Word | 93% | Comparable; some aspects better, but multimodal not used. |
| **22** | Arabic SL | CyberGlove | Words | 99.6% | Higher accuracy, but CyberGlove are less portable and require the user to wear them, which can |
| | | | | | negatively impact the overall user experience. |
| **23** | Arabic SL | RGB Camera | Words | 98% | Comparable; some aspects better, but multimodal not used. |
| **32** | ASL | EMG sensors | Alphabet , Digits | ~99% | Much higher, but uses wearable EMG different modality altogether. |
| **35** | Chinese SL | RGB Camera | Alphabet | 99.7% | Higher, but task was simpler (letter classification). |

Previous studies contain multiple significant findings that stand out:

- Leap Motion-only systems such as [14] and [19] report accuracy levels above 95%, but they often target static gestures, letters, or digits using small, controlled datasets.
- A dual Leap Motion setup in [11] achieved 92% accuracy on 100 dynamic signs, but required impractical hardware.
- A Kinect Camera such as [7] and [13] report accuracy levels above 93%, but Kinect is less portable because of its bulk, external power, and more complex setup.
- While Bird et al. [6] demonstrated strong generalization across users on BSL, our study contributes to the field by extending multimodal sign language recognition to Arabic Sign Language (ArSL), which is far less explored. Although our system achieved lower overall accuracy (78%), it outperformed single-modal baselines (Leap Motion: 59%, RGB: 70%), demonstrating the potential value of multimodal fusion in ArSL. Unlike [6], our current evaluation does not include cross-user validation; therefore, future work will aim to assess generalizability across signers.

In contrast, the current system addresses word-level gestures in ArSL, which are semantically richer and more complex than letters or numbers. Moreover, as shown in the Table 8, when comparing the same data to the single model, RGB achieved 70% accuracy, and Leap Motion achieved 59% accuracy, while multimodal achieved 78% accuracy, which means multimodal improves accuracy against (Leap Motion only vs. RGB only). Furthermore, multimodal learn features instead of model memorization because it achieved an accuracy in training was 80% and in testing was 78%.

Table 8: Accuracy Comparison Across Input Modalities

| Modality | Accuracy |
|---|---|
| **RGB** | 70% |
| **Leap Motion** | 59% |
| **Multimodal** | 78% |

This research work, although at an early stage, shows competitive results given the limitations in data and computing. The exploration of multimodal ArSL recognition is promising, especially at the word level, which is a less -tudied problem than alphabet recognition. This research contributes to the field of sign language recognition in the following ways:

- Feasibility Demonstration: It explores and demonstrates the feasibility of using a multimodal fusion approach (Leap Motion + RGB camera) for word-level ArSL gesture recognition, an area with limited prior work.
- Custom Dataset Collection: A dataset of 18 frequently used ArSL gestures was collected using a synchronized dual-sensor setup, providing valuable resources for future research.
- Prototype Implementation: The trained model was deployed in a desktop application with real-time detection and audio feedback, demonstrating an end-to-end working system.

## 5.1 Limitation

During the development and evaluation of the Arabic Sign Language (ArSL) recognition system, several challenges were encountered, primarily due to the linguistic, visual, and technical complexities associated with ArSL. One of the most notable challenges was the high variability in hand shapes and motion trajectories across signers, which introduced significant intra-class variance. This issue was further exacerbated by the morphological richness of the Arabic language and the subtle distinctions between similar signs, making fine-grained classification particularly difficult. Another critical limitation was data scarcity. Unlike datasets available for American or British Sign Languages, publicly available datasets for ArSL to be used in ML, are limited in size, diversity, and quality. This lack of sufficient and balanced training data contributed to performance degradation, particularly for underrepresented signs. Furthermore, some misclassifications were attributed to overlapping gestures or occlusion issues, where the hand obscures critical parts of the sign during capture.

To overcome these challenges, multiple enhancements were incorporated into the system. Initially, a dataset was constructed by simultaneously collecting data for 18 sign language classes using Leap Motion sensor readings alongside corresponding image captures. To address data scarcity and improve model generalization, data augmentation techniques such as random zooming and contrast adjustment were applied to synthetically expand the dataset. Furthermore, transfer learning was employed by fine-tuning pre-trained convolutional neural network (CNN) architectures, enabling more effective feature extraction from the limited available data.

### 5.1.1 Threats to Validity

In evaluating the effectiveness and generalizability of the proposed multimodal Arabic Sign Language (ArSL) recognition system, several potential.

#### Construct Validity

- **Platform and Hardware Constraints**: Prototyping on macOS (Apple M1) introduced performance bottlenecks due to the lack of CUDA support for GPU acceleration. Inference was executed on the CPU or through experimental Metal APIs. This mismatch occasionally led to increased latency and reduced responsiveness during extended use. As a mitigation strategy, future work might include model quantization, Core ML conversion, and leveraging Apple's Neural Engine.
- **Sensor Misalignment and Limitations**: Using two separate sensors (Leap Motion and RGB camera) introduces the potential for spatial or temporal misalignment, which could affect the consistency of the input data. This remains a potential source of variation in recognition accuracy.

### Internal Validity

- **Controlled Data Collection Conditions**: The dataset was collected in controlled environments, which may not reflect real-world variability in lighting, background, or user behavior. This may lead to overfitting on ideal conditions. Although this was necessary for initial prototyping, the dataset should be expanded and diversified in future phases.
- **Lack of Sensor Calibration and User Adaptation**: While the Leap Motion sensor is inherently robust in capturing 3D hand motion, dynamic adaptation strategies were not implemented to account for variations in user hand size, position, or signing style, which could impact generalizability to new users.

### External Validity

- **Limited Dataset Size and Sign Vocabulary**: The system was trained and evaluated on a dataset covering only 18 signs. Although the model successfully recognized 13 of these signs, the limited vocabulary and dataset size constrain the generalizability of the results. Dataset expansion is considered in the future work plan.
- **Absence of Real User Evaluation**: The system has not yet been tested by Deaf or Hard-of-Hearing users in real-world settings. This restricts conclusions about usability, accessibility, and user satisfaction. Usability testing will be conducted in subsequent phases of the project.

## 6 CONCLUSION

In this research, a custom dataset was created by simultaneously collecting gesture data from a Leap Motion sensor and a standard RGB camera to capture Saudi Arabic Sign Language signs. A deep learning-based recognition model was then developed and achieved an overall classification accuracy of 78%, correctly identifying 13 out of the 18 sign classes. The remaining five classes were misclassified, highlighting areas for improvement in model generalization and robustness. Finally, the trained model was saved and integrated into a desktop application to support real-time sign language detection and user interaction. However, there are ongoing challenges related to user experience and performance speed. Although the model was developed using GPU acceleration and leverages state-of-the-art architectures, saving it and using it on an M1-based device introduces performance limitations, resulting in noticeable latency during inference.

Overall, the findings suggest that integrating Leap Motion and RGB camera input may offer advantages in recognizing a broader set of ArSL gestures. However, further evaluation on larger datasets and deployment is needed to validate these results across real-world conditions. Therefore, we will focus on expanding the dataset to enhance model accuracy and generalization. Additionally, the system will be tested with end users to evaluate its real-world usability and performance, ensuring that it meets user expectations and functional requirements.

This work is one of the first to explore multimodal Arabic Sign Language recognition using Leap Motion and RGB data, and to benchmark its performance against both single-modal baselines and prior multimodal systems. While accuracy remains lower than state-of-the-art systems in BSL or ASL due to limitations mentioned above, our findings confirm the feasibility of sensor fusion for ArSL and establish a foundation for future generalizability and dataset diversity improvements.

## References

**[1]** Alzohairi, R., Alghonaim, R., Alshehri, W., & Aloqeely, S. (2018). Image based Arabic sign language recognition system. *International Journal of Advanced Computer Science and Applications*, *9*(3).

**[2]** Assaleh, K., Shanableh, T., Fanaswala, M., Amin, F., & Bajaj, H. (2010). Continuous Arabic sign language recognition in user dependent mode.

**[3]** Auti, A., Amolic, R., Bharne, S., Raina, A., & Gaikwad, D. P. (2017). Sign-Talk: Hand gesture recognition system. *International Journal of Computer Applications*, *160*(9).

**[4]** Avola, D., Bernardi, M., Cinque, L., Foresti, G. L., & Massaroni, C. (2018). Exploiting recurrent neural networks and leap motion controller for the recognition of sign language and semaphoric hand gestures. *IEEE Transactions on Multimedia*, *21*(1), 234-245.

**[5]** Bachmann, D., Weichert, F., & Rinkenauer, G. (2018). Review of three-dimensional human-computer interaction with focus on the leap motion controller. *Sensors*, *18*(7), 2194.

**[6]** Bird, J. J., Ekárt, A., & Faria, D. R. (2020). British sign language recognition via late fusion of computer vision and leap motion with transfer learning to american sign language. *Sensors*, *20*(18), 5151.

**[7]** Chai, X., Li, G., Lin, Y., Xu, Z., Tang, Y., Chen, X., & Zhou, M. (2013, April). Sign language recognition and translation with kinect. In *IEEE conf. on AFGR* (Vol. 655, p. 4).

**[8]** Cheok, M. J., Omar, Z., & Jaward, M. H. (2019). A review of hand gesture and sign language recognition techniques. *International Journal of Machine Learning and Cybernetics*, *10*, 131-153.

**[9]** Chuan, C. H., Regina, E., & Guardino, C. (2014, December). American sign language recognition using leap motion sensor. In *2014 13th International Conference on Machine Learning and Applications* (pp. 541-544). IEEE.

**[10]** DeepAI. "Convolutional Neural Network." DeepAI, https://deepai.org/machine-learning-glossary-and-terms/convolutional-neural-network

**[11]** Deriche, M., Aliyu, S. O., & Mohandes, M. (2019). An intelligent arabic sign language recognition system using a pair of LMCs with GMM based classification. *IEEE Sensors Journal*, *19*(18), 8067-8078.

**[12]** Enikeev, D., & Mustafina, S. (2020, November). Recognition of sign language using leap motion controller data. In *2020 2nd international conference on control systems, mathematical modeling, automation and energy efficiency (SUMMA)* (pp. 393-397). IEEE.

**[13]** Hisham, B., & Hamouda, A. (2019). Supervised learning classifiers for Arabic gestures recognition using Kinect V2. *SN Applied Sciences*, *1*(7), 768.

**[14]** Hisham, B., & Hamouda, A. (2017). Arabic Static and Dynamic Gestures Recognition Using Leap Motion. *J. Comput. Sci.*, *13*(8), 337-354.

**[15]** Khelil, B., Amiri, H., Chen, T., Kammüller, F., Nemli, I., & Probst, C. W. (2016, March). Hand gesture recognition using leap motion controller for recognition of arabic sign language. In *3rd International conference ACECS* (Vol. 16, pp. 233-238).

**[16]** Kudrinko, K., Flavin, E., Zhu, X., & Li, Q. (2020). Wearable sensor-based sign language recognition: A comprehensive review. *IEEE Reviews in Biomedical Engineering*, *14*, 82-97.

**[17]** Kumar, P., Saini, R., Behera, S. K., Dogra, D. P., & Roy, P. P. (2017, May). Real-time recognition of sign language gestures and air-writing using leap motion. In *2017 fifteenth IAPR international conference on machine vision applications (MVA)* (pp. 157-160). IEEE.

**[18]** Lu, C., Kozakai, M., & Jing, L. (2023). Sign Language Recognition with Multimodal Sensors and Deep Learning Methods. *Electronics*, *12*(23), 4827.

**[19]** Mohandes, M., Aliyu, S., & Deriche, M. (2014, June). Arabic sign language recognition using the leap motion controller. In *2014 IEEE 23rd International Symposium on Industrial Electronics (ISIE)* (pp. 960-965). IEEE.

**[20]** Mohandes, M., Deriche, M., & Liu, J. (2014). Image-based and sensor-based approaches to Arabic sign language recognition. *IEEE transactions on human-machine systems*, *44*(4), 551-557.

**[21]** Mohandes, M., Quadri, S. I., & Deriche, M. (2007, May). Arabic sign language recognition an image-based approach. In *21st International Conference on Advanced Information Networking and Applications Workshops (AINAW'07)* (Vol. 1, pp. 272-276). IEEE.

**[22]** Mohandes, M. A. (2013). Recognition of two-handed arabic signs using the cyberglove. *Arabian Journal for Science and Engineering*, *38*, 669-677.

**[23]** Mohandes, M., & Deriche, M. (2005). Image based Arabic sign language recognition. In *8th International Symposium on Signal Processing and its Applications, ISSPA 2005* (pp. 86-89).

**[24]** Moustafa, A. M. A., Rahim, M. M., Khattab, M. M., Zeki, A. M., Matter, S. S., Soliman, A. M., & Ahmed, A. M. (2024). Arabic Sign Language Recognition Systems: A Systematic Review. *Indian Journal of Computer Science and Engineering*, *15*, 1-18.

**[25]** Nandy, A. (2016). *Leap Motion for developers*. Apress.

**[26]** Neiva, D. H., & Zanchettin, C. (2018). Gesture recognition: A review focusing on sign language in a mobile context. *Expert Systems with Applications*, *103*, 159-183.

**[27]** Radford, A., Kim, J. W., Hallacy, C., Ramesh, A., Goh, G., Agarwal, S., ... & Sutskever, I. (2021, July). Learning transferable visual models from natural language supervision. In *International conference on machine learning* (pp. 8748-8763). PMLR.

**[28]** Saudi Society for Hearing Impairment. (n.d.). *Category*. SSHI. https://sshi.sa/category

**[29]** Satya, A. "Understanding Convolutional Neural Networks (CNN)." LearnOpenCV,https://learnopencv.com/understanding-convolutional-neural-networks-cnn/.

**[30]** Shorten, C., & Khoshgoftaar, T. M. (2019). A survey on image data augmentation for deep learning. *Journal of big data*, *6*(1), 1-48.

**[31] Simonyan, K., & Zisserman, A.** (2015). Very deep convolutional networks for large-scale image recognition. *International Conference on Learning Representations (ICLR)*. arXiv:1409.1556.

**[32]** Singh, S. K., & Chaturvedi, A. (2023). A reliable and efficient machine learning pipeline for american sign language gesture recognition using EMG sensors. *Multimedia Tools and Applications*, *82*(15), 23833-23871.

**[33]** Weiss, K., Khoshgoftaar, T. M., & Wang, D. (2016). A survey of transfer learning. *Journal of Big Data, 3*, 1–40.

**[34]** World Health Organization. (2021, September 17). Deafness and hearing loss. *WHO*.https://www.who.int/news-room/fact-sheets/detail/deafness-and-hearing-loss

**[35]** Yang, Q. (2010, June). Chinese sign language recognition based on video sequence appearance modeling. In *2010 5th IEEE Conference on Industrial Electronics and Applications* (pp. 1537-1542). IEEE.